\documentclass[twoside]{article}

\usepackage{aistats2018}
\usepackage[utf8]{inputenc} 
\usepackage[T1]{fontenc}    
\usepackage{hyperref}       
\usepackage{url}            
\usepackage{booktabs}       
\usepackage{amsfonts}       
\usepackage{nicefrac}       
\usepackage{microtype}      
\usepackage{amsmath, amssymb}        
\usepackage{setspace}

\usepackage[normalem]{ulem}

\usepackage{color}

\usepackage{caption}
\usepackage{subcaption}
\usepackage{tikz}
\usepackage{pgfplots}
\pgfplotsset{compat=1.14}

\begin{document}





\twocolumn[

\aistatstitle{Graph Convolutional Matrix Completion}

\aistatsauthor{ Rianne van den Berg \And  Thomas N.~Kipf \And  Max Welling }

\aistatsaddress{ University of Amsterdam \And  University of Amsterdam \And University of Amsterdam, CIFAR\footnotemark} ]

\footnotetext{Canadian Institute for Advanced Research}


\begin{abstract}
We consider matrix completion for recommender systems from the point of view of link prediction on graphs. Interaction data such as movie ratings can be represented by a bipartite user-item graph with labeled edges denoting observed ratings. Building on recent progress in deep learning on graph-structured data, we propose a graph auto-encoder framework based on differentiable message passing on the bipartite interaction graph. Our model shows competitive performance on standard collaborative filtering benchmarks. In settings where complimentary feature information or structured data such as a social network is available, our framework outperforms recent state-of-the-art methods. 
\end{abstract}

\section{Introduction}
With the explosive growth of e-commerce and social media platforms, recommendation algorithms have become indispensable tools for many businesses. 
Two main branches of recommender algorithms are often distinguished: content-based recommender systems \cite{pazzani2007content} and collaborative filtering models \cite{goldberg1992using}. Content-based recommender systems use content information of users and items, such as their respective occupation and genre, to predict the next purchase of a user or rating of an item. Collaborative filtering models solve the matrix completion task by taking into account the collective interaction data to predict future ratings or purchases. 

In this work, we view matrix completion as a link prediction problem on graphs: the interaction data in collaborative filtering can be represented by a bipartite graph between user and item nodes, with observed ratings/purchases represented by links. Content information can naturally be included in this framework in the form of node features. Predicting ratings then reduces to predicting labeled links in the bipartite user-item graph.

\begin{figure*}[htp!]
    \centering
    \includegraphics[width=0.75\textwidth,trim={0 0.2cm 0 0},clip]{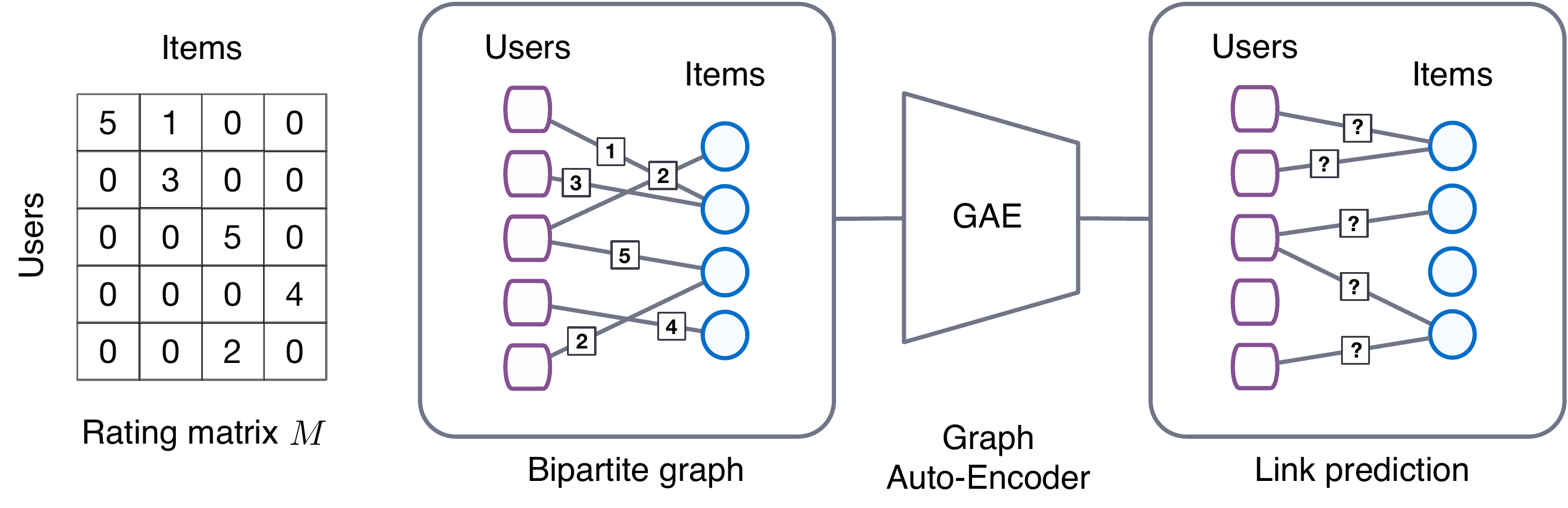}
    \caption{\textit{Left}: Rating matrix $M$ with entries that correspond to user-item interactions (ratings between 1-5) or missing observations (0). \textit{Right}: User-item interaction graph with bipartite structure. Edges correspond to interaction events, numbers on edges denote the rating a user has given to a particular item. The matrix completion task (i.e.~predictions for unobserved interactions) can be cast as a link prediction problem and modeled using an end-to-end trainable graph auto-encoder.}
    \label{fig:bipartite-graph}
\end{figure*}

We propose graph convolutional matrix completion (GC-MC): a graph-based auto-encoder framework for matrix completion, which builds on recent progress in deep learning on graphs \cite{bruna2013spectral, duvenaud2015convolutional, li2015gated, defferrard2016convolutional, kipf2016semi, tian2014learning, kipf2016variational}.
The auto-encoder produces latent features of user and item nodes through a form of message passing on the bipartite interaction graph. These latent user and item representations are used to reconstruct the rating links through a bilinear decoder. 

The benefit of formulating matrix completion as a link prediction task on a bipartite graph becomes especially apparent when recommender graphs are accompanied with structured external information such as social networks. Combining such external information with interaction data can alleviate performance bottlenecks related to the cold start problem. We demonstrate that our graph auto-encoder model efficiently combines interaction data with side information, without resorting to recurrent frameworks as in \cite{2017_Monti_arXiv}.

The paper is structured as follows: in Section \ref{section:mc_link_prediction} we introduce our graph auto-encoder model for matrix completion. Section \ref{sec:rel} discusses related work. Experimental results are shown in Section \ref{sec:exp}, and conclusion and future research directions are discussed in Section \ref{sec:conc}.

\section{Matrix completion as link prediction in bipartite graphs} 
\label{section:mc_link_prediction}
Consider a rating matrix $M$ of shape $N_u \times N_v$, where $N_u$ is the number of users and $N_v$ is the number of items. Entries $M_{ij}$ in this matrix encode either an observed rating (user $i$ rated item $j$) from a set of discrete possible rating values, or the fact that the rating is unobserved (encoded by the value $0$). See Figure \ref{fig:bipartite-graph} for an illustration. The task of matrix completion or recommendation can be seen as predicting the value of unobserved entries in $M$.

\newpage
In an equivalent picture, matrix completion or recommendation can be cast as a link prediction problem on a bipartite user-item interaction graph. More precisely, the interaction data can be represented by an undirected graph 
$G = (\mathcal{W}, \mathcal{E}, \mathcal{R})$ with entities consisting of a collection of user nodes $u_i \in \mathcal{U}$ with $i \in \{1,...,N_u\}$, and item nodes $v_{j} \in \mathcal{V}$ with $j \in \{1,...,N_v\}$, such that $\mathcal{U} \cup \mathcal{V} = \mathcal W$. The edges $(u_i, r, v_j) \in \mathcal{E}$ carry labels that represent ordinal rating levels, such as $r\in\{1, ..., R\}=\mathcal{R}$. This connection was previously explored in \cite{2013_Li_DecisionSupportSystems} and led to the development of graph-based methods for recommendation.

Previous graph-based approaches for recommender systems (see \cite{2013_Li_DecisionSupportSystems} for an overview) typically employ a multi-stage pipeline, consisting of a graph feature extraction model and a link prediction model, all of which are trained separately. Recent results, however, have shown that results can often be significantly improved by modeling graph-structured data with end-to-end learning techniques \cite{bruna2013spectral, duvenaud2015convolutional, li2015gated, niepert2016learning, defferrard2016convolutional, kipf2016semi, monti2016geometric} and specifically with graph auto-encoders \cite{tian2014learning, kipf2016variational} for unsupervised learning and link prediction. In what follows, we introduce a specific variant of graph auto-encoders for the task of recommendation.

\begin{figure*}[ht]
    \centering
    \includegraphics[width=0.8\textwidth,trim={0 0.5cm 0 0},clip]{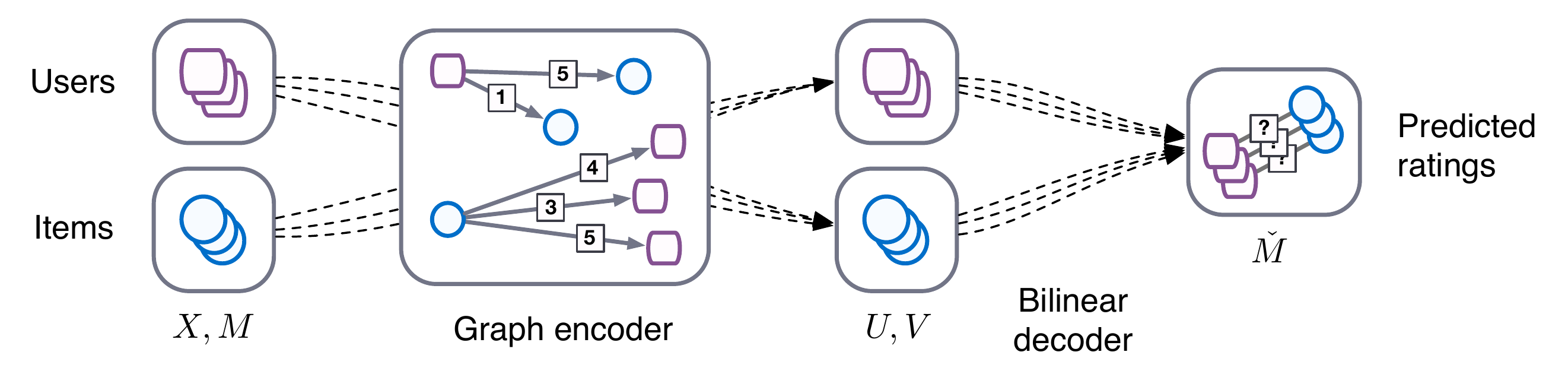}
    \caption{Schematic of a forward-pass through the GC-MC model, which is comprised of a graph convolutional encoder $[U, V] = f(X, M_1, \ldots, M_R)$ that passes and transforms messages from user to item nodes, and vice versa, followed by a bilinear decoder model that predicts entries of the (reconstructed) rating matrix $\check{M} = g(U, V)$, based on pairs of user and item embeddings.}
    \label{fig:gae}
\end{figure*}

\subsection{Graph auto-encoders}
We revisit graph auto-encoders which were originally introduced in \cite{tian2014learning, kipf2016variational} as an end-to-end model for unsupervised learning \cite{tian2014learning} and link prediction \cite{kipf2016variational} on undirected graphs. We specifically consider the setup introduced in \cite{kipf2016variational}, as it makes efficient use of (convolutional) weight sharing and allows for inclusion of side information in the form of node features. Graph auto-encoders are comprised of 1) a graph encoder model $Z = f(X, A)$, which take as input an $N\times D$ feature matrix $X$ and a graph adjacency matrix $A$, and produce an $N\times E$ node embedding matrix $Z = [z_1^T, \ldots, z_N^T]^T$, and 2) a pairwise decoder model $\check{A} = g(Z)$, which takes pairs of node embeddings $(z_i, z_j)$ and predicts respective entries $\check{A}_{ij}$ in the adjacency matrix. Note that $N$ denotes the number of nodes, $D$ the number of input features, and $E$ the embedding size.

For bipartite recommender graphs $G = (\mathcal{W}, \mathcal{E}, \mathcal{R})$, we can reformulate the encoder as $[U, V] = f(X, M_1, \ldots, M_R)$, where $M_r \in \{0,1\}^{N_u \times N_v}$ is the adjacency matrix associated with rating type $r\in\mathcal{R}$, such that $M_r$ contains $1$'s for those elements for which the original rating matrix $M$ contains observed ratings with value $r$. $U$ and $V$ are now matrices of user and item embeddings with shape $N_u\times E$ and $N_v\times E$, respectively. A single user (item) embedding takes the form of a real-valued vector $U_{i,:}$ ($V_{j,:}$) for user $i$ (item $j$). 

Analogously, we can reformulate the decoder as $\check{M} = g(U, V)$, i.e.~as a function acting on the user and item embeddings and returning a (reconstructed) rating matrix $\check{M}$ of shape $N_u \times N_v$. We can train this graph auto-encoder by minimizing the reconstruction error between the predicted ratings in $\check{M}$ and the observed ground-truth ratings in $M$. Examples of metrics for the reconstruction error are the root mean square error, or the cross entropy when treating the rating levels as different classes. 

We shall note at this point that several recent state-of-the-art models for matrix completion \cite{lee2013local, sedhain2015autorec, dziugaite2015neural, zheng2016neural} can be cast into this framework and understood as a special case of our model. An overview of these models is provided in Section \ref{sec:rel}.

\subsection{Graph convolutional encoder}
\label{sec:encoder}
In what follows, we propose a particular choice of encoder model that makes efficient use of weight sharing across locations in the graph and that assigns separate processing channels for each edge type (or rating type) $r\in\mathcal{R}$. The form of weight sharing is inspired by a recent class of convolutional neural networks that operate directly on graph-structured data \cite{bruna2013spectral, duvenaud2015convolutional, defferrard2016convolutional, kipf2016semi}, in the sense that the graph convolutional layer performs local operations that only take the first-order neighborhood of a node into account, whereby the same transformation is applied across all locations in the graph.

This type of local graph convolution can be seen as a form of message passing \cite{dai2016discriminative, gilmer2017neural}, where vector-valued messages are being passed and transformed across edges of the graph. In our case, we can assign a specific transformation for each rating level, resulting in edge-type specific messages $\mu_{j\rightarrow i,r}$ from items $j$ to users $i$ of the following form:
\begin{align}
\label{eq:messages}
\mu_{j\rightarrow i,r} = {\frac{1}{c_{ij}}}W_r x_j \, .
\end{align}
Here, $c_{ij}$ is a normalization constant, which we choose to either be $|\mathcal{N}_{i}|$ (left normalization) or $\sqrt{|\mathcal{N}_{i}||\mathcal{N}_{j}|}$ (symmetric normalization) with
 $\mathcal{N}_{i}$ denoting the set of neighbors of node $i$. $W_r$ is an edge-type specific parameter matrix and $x_j$ is the (initial) feature vector of node $j$. Messages $\mu_{i\rightarrow j,r}$ from users to items are processed in an analogous way. After the message passing step, we accumulate incoming messages at every node by summing over all neighbors $\mathcal{N}_{i,r}$ under a specific edge-type $r$, and by subsequently accumulating them into a single vector representation:
\begin{align}
\label{eq:stack}
h_i = \sigma\biggl[\mathrm{accum}\biggl(\sum_{j\in\mathcal{N}_{i,1}}\mu_{j\rightarrow i,1},\, \ldots, \sum_{j\in\mathcal{N}_{i,R}}\mu_{j\rightarrow i,R}\biggr)\biggr]  ,
\end{align}
where $\mathrm{accum}(\cdot)$ denotes an accumulation operation, such as $\mathrm{stack}(\cdot)$, i.e.~a concatenation of vectors (or matrices along their first dimension), or $\mathrm{sum}(\cdot)$, i.e.~summation of all messages. $\sigma(\cdot)$ denotes an element-wise activation function such as the $\mathrm{ReLU}(\cdot)=\mathrm{max}(0,\cdot)$. To arrive at the final embedding of user node $i$, we transform the intermediate output $h_i$ as follows:
\begin{align}
\label{eq:dense}
u_i = \sigma(W h_i) \, .
\end{align}
The item embedding $v_i$ is calculated analogously with the same parameter matrix $W$. In the presence of user- and item-specific side information we use separate parameter matrices for user and item embeddings. We will refer to (\ref{eq:stack}) as a \textit{graph convolution} layer and to (\ref{eq:dense}) as a \textit{dense} layer. Note that deeper models can be built by stacking several layers (in arbitrary combinations) with appropriate activation functions. In initial experiments, we found that stacking multiple convolutional layers did not improve performance and a simple combination of a convolutional layer followed by a dense layer worked best.

It is worth mentioning that the model demonstrated here is only one particular possible, yet relatively simple choice of an encoder, and other variations are potentially worth exploring. Instead of a simple linear message transformation, one could explore variations where $\mu_{j\rightarrow i,r}=nn(x_i, x_j, r)$ is a neural network in itself. Instead of choosing a specific normalization constant for individual messages, such as done here, one could employ some form of attention mechanism, where the individual contribution of each message is learned and determined by the model.

\subsection{Bilinear decoder}
For reconstructing links in the bipartite interaction graph we consider a bilinear decoder, and treat each rating level as a separate class. Indicating the reconstructed rating between user $i$ and item $j$ with $\check M_{ij}$, the decoder produces a probability distribution over possible rating levels through a bilinear operation followed by the application of a $\mathrm{softmax}$ function:
\begin{align}
p(\check M_{ij} = r) = \frac{e^{u_i^T Q_r v_j}}{\sum_{s \in R}e^{u_i^T Q_s v_j} } \,,
\end{align}
with $Q_r$ a trainable parameter matrix of shape $E \times E$, and $E$ the dimensionality of hidden user (item) representations $u_i$ ($v_j$).
The predicted rating is computed as 
\begin{align}
\check M_{ij} = g(u_i, v_j) = \mathbb E_{p(\check M_{ij} =r)}[r] = \sum_{r\in R} r\; p(\check M_{ij} =r) \,.
\end{align}

\subsection{Model training}
\paragraph{Loss function} During model training, we minimize the following negative log likelihood of the predicted ratings $\check{M}_{ij}$:
\begin{equation}
\label{eq:loss}
\mathcal{L} = -\sum_{i,j; \mathbf\Omega_{ij}=1}  \sum_{r=1}^R I[r=M_{ij}]  \log p(\check M_{ij} =r)\, ,
\end{equation}
with $I[k=l]=1$ when $k=l$ and zero otherwise. The matrix $\mathbf \Omega \in \{0,1\}^{N_u \times N_i}$ serves as a mask for unobserved ratings, such that ones occur for elements corresponding to observed ratings in $M$, and zeros for unobserved ratings. Hence, we only optimize over observed ratings.

\paragraph{Node dropout} 
In order for the model to generalize well to unobserved ratings, it is trained in a denoising setup by randomly dropping out all outgoing messages of a particular node, with a probability $p_{\mathrm{dropout}}$, which we will refer to as \textit{node dropout}. Messages are rescaled after dropout as in \cite{srivastava2014dropout}. In initial experiments we found that node dropout was more efficient in regularizing than message dropout. In the latter case individual outgoing messages are dropped out independently, making embeddings more robust against the presence or absence of single edges. In contrast, node dropout  also causes embeddings to be more independent of particular user or item influences. We furthermore also apply regular dropout \cite{srivastava2014dropout} to the hidden layer units (\ref{eq:dense}).

\paragraph{Mini-batching}
We introduce mini-batching by sampling contributions to the loss function in Eq.~\eqref{eq:loss} from different observed ratings. That is, we sample only a fixed number of contributions from the sum over user and item pairs. By only considering a fixed number of contributions to the loss function, we can remove respective rows of users and items in $M_1, ..., M_R$ in Eq.~\eqref{eq:encoder-vec} that do not appear in the current batch. This serves both as an effective means of regularization, and reduces the memory requirement to train the model, which is necessary to fit the full model for MovieLens-10M into GPU memory. 
We experimentally verified that training with mini-batches and full batches leads to comparable results for the MovieLens-1M dataset while adjusting for regularization parameters. For all datasets except for the MovieLens-10M, we opt for full-batch training since it leads to faster convergence than training with mini-batches in this particular setting.

\subsection{Vectorized implementation}
In practice, we can use efficient sparse matrix multiplications, with complexity linear in the number of edges, i.e.~$\mathcal{O}(|\mathcal{E}|)$, to implement the graph auto-encoder model. The graph convolutional encoder (Eq.~\ref{eq:dense}), for example in the case of left normalization, can be vectorized as follows:
\begin{align}
\label{eq:encoder-vec}
{U\brack V} = f(X, M_1, \ldots, M_R) = \sigma\biggl({H_u\brack H_v} W^T \biggr)\,\, ,\\ \text{with} \quad {H_u\brack H_v} = \sigma\biggl(\sum_{r=1}^R D^{-1}\mathcal M_rX W_r^T\biggr) \, , \label{eq:encoder-vec-b}\\
 \text{and} \quad \mathcal M_r = \begin{pmatrix}
0 & M_r\\
M_r^T & 0
\end{pmatrix} \, . \label{eq:encoder-vec-c}
\end{align}
The summation in \eqref{eq:encoder-vec-b} can be replaced with concatenation, similar to \eqref{eq:stack}.
In this case $D$ denotes the diagonal node degree matrix with nonzero elements $D_{ii}=|\mathcal{N}_i|$. Vectorization for an encoder with a symmetric normalization, as well as vectorization of the bilinear decoder, follows in an analogous manner. Note that it is only necessary to evaluate observed elements in $\check{M}$, given by the mask $\mathbf{\Omega}$ in Eq.~\ref{eq:loss}.

\subsection{Input feature representation and side information}
\label{subsection:features}
Features containing information for each node, such as content information, can in principle be injected into the graph encoder directly at the input-level (i.e.~in the form of an input feature matrix $X$). However, when the content information does not contain enough information to distinguish different users (or items) and their interests, feeding the content information directly into the graph convolution layer leads to a severe bottleneck of information flow. In such cases, 
one can include side information in the form of user and item feature vectors $x_i^f$ (for node $i$) via a separate processing channel directly into the the dense hidden layer:
\begin{align}
\label{eq:dense-feat}
u_i = \sigma(W h_i + W_{2}^{f} f_i) \,\,\,  \,\,\, \text{with} \quad f_i = \sigma(W_{1}^{f} x^{f}_{i} + b) \, ,
\end{align}
where $W_1^f$ and $W_2^f$ are trainable weight matrices, and $b$ is a bias. The weight matrices and bias vector are different for users and items. The input feature matrix $X=[x_1^T, \ldots, x_N^T]^T$ containing the node features for the graph convolution layer is then chosen as an identity matrix, with a unique one-hot vector for every node in the graph. For the datasets considered in this paper, the user (item) content information is of limited size, and we thus opt to include this as side information while using Eq. \eqref{eq:dense-feat}.

In \cite{strub2016hybrid}, Strub et al.~propose to include content information along similar lines, although in their case the proposed model is strictly user- or item-based, and thus only supports side information for either users or items.

Note that side information does not necessarily need to come in the form of per-node feature vectors, but can also be provided in the form of, e.g., graph-structured, natural language, or image data. In this case, the dense layer in (\ref{eq:dense-feat}) is replaced by an appropriate differentiable module, such as a recurrent neural network, a convolutional neural network, or another graph convolutional network.

\subsection{Weight sharing}
\label{sec:ord}
In the collaborative filtering setting with one-hot vectors as input, the columns of the weight matrices $W_r$ play the role of latent factors for each separate node for one specific rating value $r$. These latent factors are passed onto connected user or item nodes through message passing.
However, not all users and items necessarily have an equal number of ratings for each rating level. 
This results in certain columns of $W_r$ to be optimized significantly less frequently than others. Therefore, some form of weight sharing between the matrices $W_r$ for different $r$ is desirable to alleviate this optimization problem. 
Following \cite{zheng2016neural}, we therefore implement the following weight sharing setup:
\begin{equation}
W_r = \sum_{s=1}^r T_s \, .
\end{equation}
We will refer to this type of weight sharing as ordinal weight sharing due to the increasing number of weight matrices included for higher rating levels. 

As an effective means of regularization of the pairwise bilinear decoder, we resort to weight sharing in the form of a linear combination of a set of basis weight matrices $P_s$:
\begin{align}
Q_r = \sum_{s=1}^{n_b} a_{rs} P_s\, ,
\end{align}
with $s\in (1,...,n_b)$ and $n_b$ being the number of basis weight matrices. Here, $a_{rs}$ are the learnable coefficients that determine the linear combination for each decoder weight matrix $Q_r$. Note that in order to avoid overfitting and to reduce the number of parameters, the number of basis weight matrices $n_b$ should naturally be lower than the number of rating levels. 

\section{Related work}
\label{sec:rel}
\paragraph{Auto-encoders} User- or item-based auto-encoders \cite{sedhain2015autorec, zheng2016neural,strub2016hybrid} are a recent class of state-of-the-art collaborative filtering models that can be seen as a special case of our graph auto-encoder model, where only either user or item embeddings are considered in the encoder. AutoRec by Sedhain et al. \cite{sedhain2015autorec} is the first such model, where the user's (or item's) partially observed rating vector is projected onto a latent space through an encoder layer, and reconstructed using a decoder layer with mean squared reconstruction error loss. 

The CF-NADE algorithm by Zheng et al. \cite{zheng2016neural} can be considered as a special case of the above auto-encoder architecture. In the user-based setting, messages are only passed from items to users, and in the item-based case the reverse holds. Note that in contrast to our model, unrated items are assigned a default rating of $3$ in the encoder, thereby creating a fully-connected interaction graph. CF-NADE imposes a random ordering on nodes, and splits incoming messages into two sets via a random cut, only one of which is kept. This model can therefore be seen as a denoising auto-encoder, where part of the input space is dropped out at random in every iteration.

\paragraph{Factorization models}
Many of the most popular collaborative filtering algorithms fall into the class of matrix factorization (MF) models. Methods of this sort assume the rating matrix to be well approximated by a low rank matrix: $M \approx U V^T$, with $U \in \mathbb{R}^{N_u \times k}$ and $V \in \mathbb{R}^{N_i \times k}$, with $k \ll N_u, N_i$. The rows of $U$ and $V$ can be seen as latent feature representations of users and items, representing an encoding for their interests through their rating pattern. Probabilistic matrix factorization (PMF) by Salakhutdinov et al. \cite{salakhutdinov2007probabilistic} assumes that the ratings contained in $M$ are independent stochastic variables with Gaussian noise. Optimization of the maximum likelihood then leads one to minimize the mean squared error between the observed entries in $M$ and the reconstructed ratings in $U V^T$. BiasedMF by Koren et al. \cite{Koren2009matrix} improves upon PMF by incorporating a user and item specific bias, as well as a global bias. Neural network matrix factorization (NNMF) \cite{dziugaite2015neural} extends the MF approach by passing the latent user and item features through a feed forward neural network. Local low rank matrix approximation by Lee et al. \cite{lee2013local}, introduces the idea of reconstructing rating matrix entries using different (entry dependent) combinations of low rank approximations.

\paragraph{Matrix completion with side information}
In matrix completion (MC) \cite{candes2012exact}, the objective is to approximate the rating matrix with a low-rank rating matrix. Rank minimization, however, is an intractable problem, and Candes \& Recht \cite{candes2012exact} replaced the rank minimization with a minimization of the nuclear norm (the sum of the singular values of a matrix), turning the objective function into a tractable convex one. Inductive matrix completion (IMC) by Jain \& Dhillon, 2013 and Xu et al., 2013 
incorporates content information of users and items in feature vectors and approximates the observed elements of the rating matrix as $M_{ij} = x^T_i U V^T y_j$, with $x_i$ and $y_j$ representing the feature vector of user $i$ and item $j$ respectively. 

The geometric matrix completion (GMC) model proposed by Kalofolias et al. in 2014 \cite{kalofolias2014matrix} introduces a regularization of the MC model by adding side information in the form of user and item graphs. In \cite{GralsRaoNIPS2015}, a more efficient  alternating least squares optimization optimization method (GRALS) is introduced to the graph-regularized matrix completion problem. Most recently, Monti et al.~\cite{2017_Monti_arXiv} suggested to incorporate graph-based side information in matrix completion via the use of convolutional neural networks on graphs, combined with a recurrent neural network to model the dynamic rating generation process. Their work is different from ours, in that we model the rating graph directly using a graph convolutional encoder/decoder approach that predicts unseen ratings in a single, non-iterative step.

\section{Experiments}
\label{sec:exp}
We evaluate our model on a number of common collaborative filtering benchmark datasets: MovieLens\footnote{\url{https://grouplens.org/datasets/movielens/}} (100K, 1M, and 10M), Flixster, Douban, and YahooMusic. The datasets consist of user ratings for items (such as movies) and optionally incorporate additional user/item information in the form of features. For Flixster, Douban, and YahooMusic we use preprocessed subsets of these datasets provided by \cite{2017_Monti_arXiv}\footnote{\url{https://github.com/fmonti/mgcnn}}. These datasets contain sub-graphs of 3000 users and 3000 items and their respective user-user and item-item interaction graphs (if available). Dataset statistics are summarized in Table \ref{table:datasets}. 

\begin{table*}[ht]
\centering
\begin{tabular}{lrrrrrr}
\toprule
\textbf{Dataset} & \textbf{Users} & \textbf{Items} & \textbf{Features} & \textbf{Ratings} & \textbf{Density} & \textbf{Rating levels} 
\\[0.05em]\midrule \\[-0.8em]
Flixster  & 3,000  & 3,000 & Users/Items & 26,173 & 0.0029 & 0.5, 1, \ldots, 5\\
Douban  & 3,000  & 3,000 & Users & 136,891 & 0.0152 & 1, 2, \ldots, 5\\
YahooMusic  & 3,000  & 3,000 & Items  & 5,335 & 0.0006 & 1, 2, \ldots, 100\\
MovieLens 100K (ML-100K)  & 943  & 1,682  & Users/Items & 100,000 & 0.0630 & 1, 2, \ldots, 5\\
MovieLens 1M (ML-1M)    & 6,040  & 3,706 & ---  & 1,000,209 & 0.0447& 1, 2, \ldots, 5 \\
MovieLens 10M (ML-10M)   & 69,878 & 10,677 & --- & 10,000,054 & 0.0134& 0.5, 1, \ldots, 5  \\ \hline 
\end{tabular}
\caption{Number of users, items and ratings for each of the MovieLens datasets used in our experiments. We further indicate rating density and rating levels.}
\label{table:datasets}
\end{table*}
For all experiments, we choose from the following settings based on validation performance: accumulation function ($\mathrm{stack}$ vs.~$\mathrm{sum}$), whether to use ordinal weight sharing in the encoder, left vs.~symmetric normalization, and dropout rate $p_{\text{dropout}}\in\{0.3, 0.4, 0.5, 0.6, 0.7, 0.8\}$. Unless otherwise noted, we use a Adam \cite{kingma2014adam} with a learning rate of $0.01$, weight sharing in the decoder with 2 basis weight matrices, and layer sizes of $500$ and $75$ for the graph convolution (with $\mathrm{ReLU}$) and dense layer (no activation function), respectively. We evaluate our model on the held out test sets using an exponential moving average of the learned model parameters with a decay factor set to $0.995$.
\paragraph{MovieLens 100K}
For this task, we compare against matrix completion baselines that make use of side information in the form of user/item features. We report performance on the canonical $\mathrm{u1.base}$/$\mathrm{u1.test}$ train/test split. Hyperparameters are optimized on a 80/20 train/validation split of the original training set. Side information is present both for users (e.g. age, gender, and occupation) and movies (genres). Following Rao et al.~\cite{GralsRaoNIPS2015}, we map the additional information onto feature vectors for users and movies, and compare the performance of our model with (GC-MC+Feat) and without the inclusion of these features (GC-MC) . Note that GMC \cite{kalofolias2014matrix}, GRALS \cite{GralsRaoNIPS2015} and sRGCNN \cite{2017_Monti_arXiv} represent user/item features via a k-nearest-neighbor graph. We use stacking as an accumulation function in the graph convolution layer in Eq. \eqref{eq:stack}, set dropout equal to $0.7$, and use left normalization. GC-MC+Feat uses 10 hidden units for the dense side information layer (with $\mathrm{ReLU}$ activation) as described in Eq.~\ref{eq:dense-feat}. We train both models for 1,000 full-batch epochs. We report RMSE scores averaged over 5 runs with random initializations\footnote{Standard error less than $0.001$.}. Results are summarized in Table \ref{tab:ml_100k_scores}.
\paragraph{MovieLens 1M and 10M}
We compare against current state-of-the-art collaborative filtering algorithms, such as AutoRec \cite{sedhain2015autorec}, LLorma \cite{lee2013local}, and CF-NADE \cite{zheng2016neural}. Results are reported as averages over the same five 90/10 training/test set splits as in \cite{zheng2016neural} and summarized in Table \ref{tab:cfnade_baseline}. Model choices are validated on an internal 95/5 split of the training set. For ML-1M we use accumulate messages through summation in Eq.  \eqref{eq:stack}, use a dropout rate of $0.7$, and symmetric normalization. 
As ML-10M has twice the number of rating classes, we use twice the number of basis function matrices in the decoder. Furthermore, we use stacking accumulation, a dropout of $0.3$ and symmetric normalization.
We train for 3,500 full-batch epochs, and 18,000 mini-batch iterations (20 epochs with batch size 10,000) on the ML-1M and ML-10M dataset, respectively.  

\paragraph{Flixster, Douban and YahooMusic} 
These datasets contain user and item side information in the form of graphs. We integrate this graph-based side information into our framework by using the adjacency vector (normalized by degree) as a feature vector for the respective user/item. For a single dense feature embedding layer, this is equivalent to performing a graph convolution akin to \cite{kipf2016semi} on the user-user or item-item graph.\footnote{With a row-normalized adjacency matrix instead of the symmetric normalization from \cite{kipf2016semi}. Both perform similarly in practice.} We use a dropout rate of $0.7$, and 64 hidden units for the dense side information layer (with $\mathrm{ReLU}$ activation) as described in Eq.~\ref{eq:dense-feat}. We use a left normalization, and messages in the graph convolution layer are accumulated by concatenation (as opposed to summation). All models are trained for $200$ epochs. For hyperparameter selection, we set aside a separate 80/20 train/validation split from the original training set in \cite{2017_Monti_arXiv}. For final model evaluation, we train on the full original training set from \cite{2017_Monti_arXiv} and report test set performance. Results are summarized in Table \ref{tab:monti}.

\begin{table}[ht]
\centering
\begin{tabular}{l r}
\toprule
\textbf{Model} & \textbf{ML-100K + Feat} 
\\[0.05em]\midrule \\[-0.8em]
MC \cite{candes2012exact} & $0.973$ \\
IMC \cite{jain2013provable, xu2013speedup}& $1.653$ \\
GMC \cite{kalofolias2014matrix} & $0.996$ \\
GRALS \cite{GralsRaoNIPS2015} & $0.945$\\
sRGCNN \cite{2017_Monti_arXiv} & $0.929$\\
GC-MC (Ours) & $0.910$ \\
GC-MC+Feat & $\mathbf{0.905}$ \\
\bottomrule
\end{tabular}
\caption{RMSE scores\protect\footnotemark\, for the MovieLens 100K task with side information on a canonical 80/20 training/test set split. Side information is either presented as a nearest-neighbor graph in user/item feature space or as raw feature  vectors. Baseline numbers are taken from \cite{2017_Monti_arXiv}.}\label{tab:ml_100k_scores}
\end{table}
\footnotetext{Results for our model slightly differ from the previous version of this paper, as we chose a different method for weight sharing in the encoder for consistency across experiments. See Section \ref{sec:encoder} for details.}

\begin{table}
\centering
\resizebox{\columnwidth}{!}{%
\begin{tabular}{l r r r}
\toprule
\textbf{Model} & \textbf{Flixster} & \textbf{Douban} & \textbf{YahooMusic} \\[0.05em]\midrule \\[-0.8em]
GRALS  & $1.313/1.245$ & $0.833$ & $38.0$\\
sRGCNN & $1.179/0.926$ & $0.801$ & $22.4$\\
GC-MC  & $\mathbf{0.941}/\mathbf{0.917}$ & $\mathbf{0.734}$ & $\mathbf{20.5}$\\
\bottomrule
\end{tabular}
}
\caption{Average RMSE test set scores for 5 runs on Flixster, Douban, and YahooMusic, all of which include side information in the form of user and/or item graphs. We replicate the benchmark setting as in \cite{2017_Monti_arXiv}. For Flixster, we show results for both user/item graphs (right number) and user graph only (left number). Baseline numbers are taken from \cite{2017_Monti_arXiv}.}\label{tab:monti} 
\end{table}
\begin{table}
\centering
\begin{tabular}{l r r}
\toprule
\textbf{Model} & \textbf{ML-1M} & \textbf{ML-10M} \\[0.05em]\midrule \\[-0.8em]
PMF \cite{salakhutdinov2007probabilistic} & $0.883$ & --\\
I-RBM \cite{salakhutdinov2007restricted} & $0.854$ & $0.825$\\
BiasMF \cite{Koren2009matrix} & $0.845$ & $0.803$\\
NNMF \cite{dziugaite2015neural} & $0.843$ & --\\
LLORMA-Local \cite{lee2013local} & $0.833$ & $0.782$ \\
I-AUTOREC \cite{sedhain2015autorec} & $0.831$ & $0.782$ \\
CF-NADE \cite{zheng2016neural} & $\mathbf{0.829}$ & $\mathbf{0.771}$ \\
GC-MC (Ours) & $0.832$ & $0.777$ \\ 
\bottomrule
\end{tabular}
\caption{Comparison of average test RMSE scores on five 90/10 training/test set splits (as in \cite{zheng2016neural}) without the use of side information. Baseline scores are taken from \cite{zheng2016neural}. For CF-NADE, we report the best-performing model variant.}\label{tab:cfnade_baseline} 
\end{table}

\paragraph{Cold-start analysis}
\begin{figure}[t!]
    \centering
    \includegraphics[width=\linewidth,trim={0 0.4cm 0 0},clip]{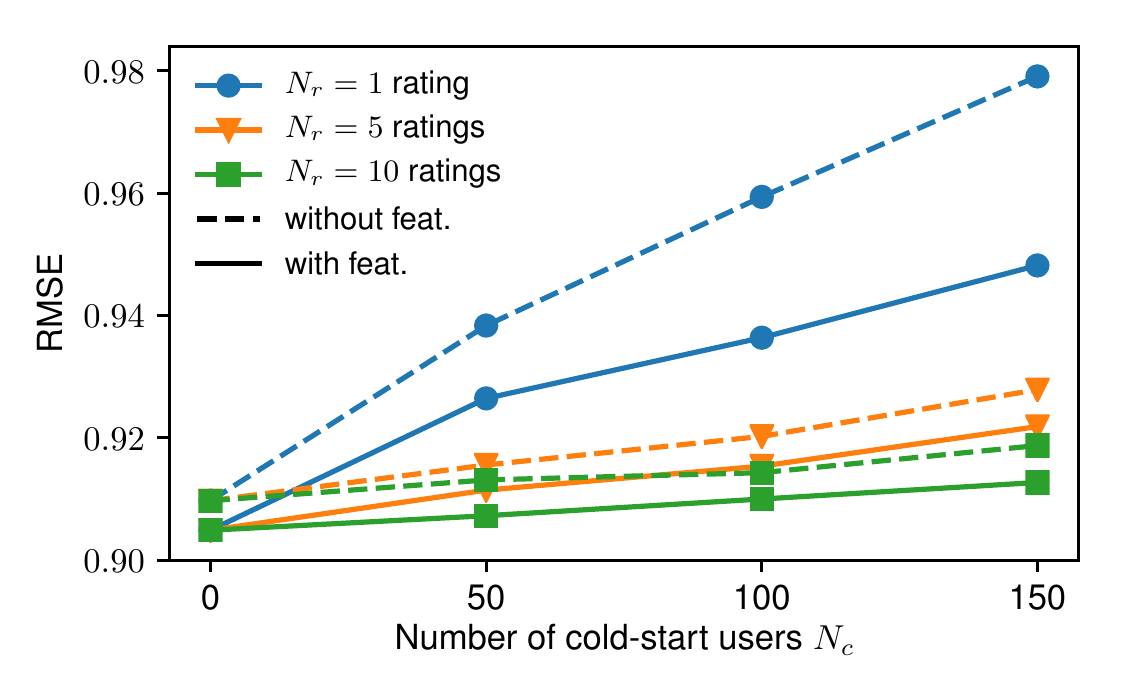}
    \caption{Cold-start analysis for ML-100K. Test set RMSE (average over 5 runs with random initialization) for various settings, where only a small number of ratings $N_r$ is kept for a certain number of cold-start users $N_c$ during training. Standard error is below $0.001$ and therefore not shown. Dashed and solid lines denote experiments without and with side information, respectively.}
    \label{fig:cold-start}
\end{figure}
To gain insight into how the GC-MC model makes use of side information, we study the performance of our model in the presence of users with only very few ratings (cold-start users). We adapt the ML-100K benchmark dataset, so that for a fixed number of cold-start users $N_c$ all ratings except for a minimum number $N_r$ are removed from the training set (chosen at random with a fixed seed across experiments). Note that ML-100K in its original form contains only users with at least 20 ratings.

We analyze model performance for $N_r\in\{1,5,10\}$ and $N_c\in\{0,50, 100,150\}$, both with and without using user/item features as side information (see Figure \ref{fig:cold-start}). Hyperparameters and test set are chosen as before, i.e.~we report RMSE on the complete canonical test set split. The benefit of incorporating side information, such as user and item features, is especially pronounced in the presence of many users with only a single rating.

\paragraph{Discussion}
On the ML-100K task with side information, our model outperforms related methods by a significant margin. Remarkably, this is even the case without the use of side information. Most related to our method is sRGCNN by Monti et al.~\cite{2017_Monti_arXiv} that uses graph convolutions on the nearest-neighbor graphs of users and items, and learns representations in an iterative manner using recurrent neural networks. Our results demonstrate that a direct estimation of the rating matrix from learned user/item representations using a simple decoder model can be more effective, while being computationally more efficient.

Our results on ML-1M and ML-10M demonstrate that it is possible to scale our method to larger datasets, putting it into the vicinity of recent state-of-the-art collaborative filtering user- or item-based methods in terms of predictive performance. At this point, it is important to note that several techniques introduced in CF-NADE \cite{zheng2016neural}, such as layer-specific learning rates, a special ordinal loss function, and the auto-regressive modeling of ratings, can be seen as orthogonal to our approach and can be used in conjunction with our framework.

For the Flixster, Douban, and YahooMusic datasets our model achieves state-of-the-art results, while using a single hyperparameter setting across all three datasets.

\section{Conclusions}
\label{sec:conc}

In this work, we have introduced graph convolutional matrix completion (GC-MC): a graph auto-encoder framework for the matrix completion task in recommender systems. The encoder contains a graph convolution layer that constructs user and item embeddings through message passing on the bipartite user-item interaction graph. Combined with a bilinear decoder, new ratings are predicted in the form of labeled edges.  

The graph auto-encoder framework naturally generalizes to include side information for both users and items.
In this setting, our proposed model outperforms recent related methods by a large margin, as demonstrated on a number of benchmark datasets with feature- and graph-based side information. We further show that our model can be trained on larger scale datasets through stochastic mini-batching. In this setting, our model achieves results that are competitive with recent state-of-the-art collaborative filtering.

In future work, we wish to extend this model to large-scale multi-modal data (comprised of text, images, and other graph-based information), such as present in many realisitic recommendation platforms. In such a setting, the GC-MC model can be combined with recurrent (for text) or convolutional neural networks (for images). To address scalability, it is necessary to develop efficient approximate schemes, such as subsampling local neighborhoods \cite{hamilton2017inductive}. Finally, attention mechanisms \cite{bahdanau2014neural} provide a promising future avenue for extending this class of models.

\subsection*{Acknowledgments}
We would like to thank Jakub Tomczak, Christos Louizos, Karen Ullrich and Peter Bloem for helpful discussions and comments.
This project is supported by the SAP Innovation
Center Network.

\begin{spacing}{0.9}
{\small
\bibliography{bibliography}

\begin{thebibliography}{10}

\bibitem{bahdanau2014neural}
Dzmitry Bahdanau, Kyunghyun Cho, and Yoshua Bengio.
\newblock Neural machine translation by jointly learning to align and
  translate.
\newblock {\em arXiv preprint arXiv:1409.0473}, 2014.

\bibitem{bruna2013spectral}
Joan Bruna, Wojciech Zaremba, Arthur Szlam, and Yann LeCun.
\newblock Spectral networks and locally connected networks on graphs.
\newblock {\em arXiv preprint arXiv:1312.6203}, 2013.

\bibitem{candes2012exact}
Emmanuel Candes and Benjamin Recht.
\newblock Exact matrix completion via convex optimization.
\newblock {\em Communications of the ACM}, 55(6):111--119, 2012.

\bibitem{dai2016discriminative}
Hanjun Dai, Bo~Dai, and Le~Song.
\newblock Discriminative embeddings of latent variable models for structured
  data.
\newblock In {\em International Conference on Machine Learning (ICML)}, 2016.

\bibitem{defferrard2016convolutional}
Micha{\"e}l Defferrard, Xavier Bresson, and Pierre Vandergheynst.
\newblock Convolutional neural networks on graphs with fast localized spectral
  filtering.
\newblock In {\em Advances in Neural Information Processing Systems}, pages
  3837--3845, 2016.

\bibitem{duvenaud2015convolutional}
David~K. Duvenaud, Dougal Maclaurin, Jorge Iparraguirre, Rafael Bombarell,
  Timothy Hirzel, Al{\'a}n Aspuru-Guzik, and Ryan~P. Adams.
\newblock Convolutional networks on graphs for learning molecular fingerprints.
\newblock In {\em Advances in neural information processing systems (NIPS)},
  pages 2224--2232, 2015.

\bibitem{dziugaite2015neural}
Gintare~Karolina Dziugaite and Daniel~M Roy.
\newblock Neural network matrix factorization.
\newblock {\em arXiv preprint arXiv:1511.06443}, 2015.

\bibitem{gilmer2017neural}
Justin Gilmer, Samuel~S. Schoenholz, Patrick~F. Riley, Oriol Vinyals, and
  George~E. Dahl.
\newblock Neural message passing for quantum chemistry.

\bibitem{goldberg1992using}
David Goldberg, David Nichols, Brian~M. Oki, and Douglas Terry.
\newblock Using collaborative filtering to weave an information tapestry.
\newblock {\em Communications of the ACM}, 35(12):61--70, 1992.

\bibitem{hamilton2017inductive}
William~L. Hamilton, Rex Ying, and Jure Leskovec.
\newblock Inductive representation learning on large graphs.
\newblock {\em arXiv preprint arXiv:1706.02216}, 2017.

\bibitem{jain2013provable}
Prateek Jain and Inderjit~S Dhillon.
\newblock Provable inductive matrix completion.
\newblock {\em arXiv preprint arXiv:1306.0626}, 2013.

\bibitem{kalofolias2014matrix}
Vassilis Kalofolias, Xavier Bresson, Michael Bronstein, and Pierre
  Vandergheynst.
\newblock Matrix completion on graphs.
\newblock {\em arXiv preprint arXiv:1408.1717}, 2014.

\bibitem{kingma2014adam}
Diederik Kingma and Jimmy Ba.
\newblock Adam: A method for stochastic optimization.
\newblock {\em arXiv preprint arXiv:1412.6980}, 2014.

\bibitem{kipf2016variational}
Thomas~N. Kipf and Max Welling.
\newblock Variational graph auto-encoders.
\newblock {\em NIPS Bayesian Deep Learning Workshop}, 2016.

\bibitem{kipf2016semi}
Thomas~N. Kipf and Max Welling.
\newblock Semi-supervised classification with graph convolutional networks.
\newblock {\em ICLR}, 2017.

\bibitem{Koren2009matrix}
Yehuda Koren, Robert Bell, and Chris Volinsky.
\newblock Matrix factorization techniques for recommender systems.
\newblock {\em Computer}, 42(8):30--37, August 2009.

\bibitem{lee2013local}
Joonseok Lee, Seungyeon Kim, Guy Lebanon, and Yoram Singer.
\newblock Local low-rank matrix approximation.
\newblock In Sanjoy Dasgupta and David McAllester, editors, {\em Proceedings of
  the 30th International Conference on Machine Learning (ICML)}, volume~28 of
  {\em Proceedings of Machine Learning Research}, pages 82--90, Atlanta,
  Georgia, USA, 17--19 Jun 2013. PMLR.

\bibitem{2013_Li_DecisionSupportSystems}
Xin Li and Hsinchun Chen.
\newblock Recommendation as link prediction in bipartite graphs: A graph
  kernel-based machine learning approach.
\newblock {\em Decision Support Systems}, 54(2):880 -- 890, 2013.

\bibitem{li2015gated}
Yujia Li, Daniel Tarlow, Marc Brockschmidt, and Richard Zemel.
\newblock Gated graph sequence neural networks.
\newblock {\em ICLR}, 2016.

\bibitem{salakhutdinov2007probabilistic}
Andriy Mnih and Ruslan~R. Salakhutdinov.
\newblock Probabilistic matrix factorization.
\newblock In {\em Advances in neural information processing systems}, pages
  1257--1264, 2008.

\bibitem{monti2016geometric}
Federico Monti, Davide Boscaini, Jonathan Masci, Emanuele Rodol{\`a}, Jan
  Svoboda, and Michael~M Bronstein.
\newblock Geometric deep learning on graphs and manifolds using mixture model
  cnns.
\newblock {\em CVPR}, 2017.

\bibitem{2017_Monti_arXiv}
Federico Monti, Michael~M. Bronstein, and Xavier Bresson.
\newblock Geometric matrix completion with recurrent multi-graph neural
  networks.
\newblock {\em NIPS}, 2017.

\bibitem{niepert2016learning}
Mathias Niepert, Mohamed Ahmed, and Konstantin Kutzkov.
\newblock Learning convolutional neural networks for graphs.
\newblock In {\em Proceedings of the 33rd annual international conference on
  machine learning. ACM}, 2016.

\bibitem{pazzani2007content}
Michael~J Pazzani and Daniel Billsus.
\newblock Content-based recommendation systems.
\newblock In {\em The adaptive web}, pages 325--341. Springer, 2007.

\bibitem{GralsRaoNIPS2015}
Nikhil Rao, Hsiang-Fu Yu, Pradeep~K. Ravikumar, and Inderjit~S. Dhillon.
\newblock Collaborative filtering with graph information: Consistency and
  scalable methods.
\newblock In C.~Cortes, N.~D. Lawrence, D.~D. Lee, M.~Sugiyama, and R.~Garnett,
  editors, {\em Advances in Neural Information Processing Systems 28}, pages
  2107--2115. Curran Associates, Inc., 2015.

\bibitem{salakhutdinov2007restricted}
Ruslan Salakhutdinov, Andriy Mnih, and Geoffrey Hinton.
\newblock Restricted boltzmann machines for collaborative filtering.
\newblock In {\em Proceedings of the 24th international conference on Machine
  learning}, pages 791--798. ACM, 2007.

\bibitem{sedhain2015autorec}
Suvash Sedhain, Aditya~Krishna Menon, Scott Sanner, and Lexing Xie.
\newblock Autorec: Autoencoders meet collaborative filtering.
\newblock In {\em Proceedings of the 24th International Conference on World
  Wide Web}, pages 111--112. ACM, 2015.

\bibitem{srivastava2014dropout}
Nitish Srivastava, Geoffrey~E. Hinton, Alex Krizhevsky, Ilya Sutskever, and
  Ruslan Salakhutdinov.
\newblock Dropout: a simple way to prevent neural networks from overfitting.
\newblock {\em Journal of Machine Learning Research}, 15(1):1929--1958, 2014.

\bibitem{strub2016hybrid}
Florian Strub, Romaric Gaudel, and J{\'e}r{\'e}mie Mary.
\newblock Hybrid recommender system based on autoencoders.
\newblock In {\em Proceedings of the 1st Workshop on Deep Learning for
  Recommender Systems}, DLRS 2016, pages 11--16, New York, NY, USA, 2016. ACM.

\bibitem{tian2014learning}
Fei Tian, Bin Gao, Qing Cui, Enhong Chen, and Tie-Yan Liu.
\newblock Learning deep representations for graph clustering.
\newblock In {\em AAAI}, pages 1293--1299, 2014.

\bibitem{xu2013speedup}
Miao Xu, Rong Jin, and Zhi-Hua Zhou.
\newblock Speedup matrix completion with side information: Application to
  multi-label learning.
\newblock In {\em Advances in Neural Information Processing Systems}, pages
  2301--2309, 2013.

\bibitem{zheng2016neural}
Yin Zheng, Bangsheng Tang, Wenkui Ding, and Hanning Zhou.
\newblock A neural autoregressive approach to collaborative filtering.
\newblock In {\em Proceedings of the 33nd International Conference on Machine
  Learning}, pages 764--773, 2016.

\end{thebibliography}
\bibliographystyle{plain}
}
\end{spacing}

\end{document}